\documentclass[letterpaper, 10 pt, conference]{ieeeconf}  

\IEEEoverridecommandlockouts                              

\usepackage{epsfig} 
\usepackage{times} 
\usepackage{amsmath} 
\usepackage{amssymb}  
\usepackage{graphicx}
\graphicspath{{images/}}
\usepackage{blindtext}
\usepackage{subfiles} 
\usepackage{dsfont}
\usepackage{booktabs}
\usepackage{subcaption}
\usepackage{gensymb}
\usepackage{xcolor}

\DeclareMathOperator*{\argmax}{argmax}

\usepackage{my_commands}

\title{\LARGE \bf
\name: Visual Base Pose Learning for Mobile Manipulation using Equivariant TransporterNet and GNNs
}

\author{Lakshadeep Naik$^{1}$, Adam Fischer$^{1}$, Daniel Duberg$^{1}$, and Danica Kragic$^{1}$
\thanks{This work was supported by the Swedish Research Council, Knut and Alice Wallenberg Foundation, and Horizon2020 ERC BIRD}
\thanks{All authors are with the division of Robotics, Perception, and Learning (RPL), KTH Royal Institute of Technology, Stockholm, Sweden
        {\tt\small laksh,adamfi,dduberg,dani@kth.se}}%
}

\begin{document}

\maketitle
\thispagestyle{empty}
\pagestyle{empty}

\begin{abstract}
 In Mobile Manipulation, selecting an optimal mobile base pose is essential for successful object grasping. Previous works have addressed this problem either through classical planning methods 
 or by learning state-based policies. They assume access to reliable state information, such as the precise object poses and environment models. In this work, we study base pose planning directly from top-down orthographic projections of the scene, which provide a global overview of the scene while preserving spatial structure. We propose \name, a learning-based method for base pose selection using such top-down orthographic projections. We use equivariant TransporterNet to exploit spatial symmetries and efficiently learn candidate base poses for grasping. Further, we use graph neural networks to represent a varying number of candidate base poses and use Reinforcement Learning to determine the optimal base pose among them. We show that \name~can produce comparable solutions to the classical methods in significantly less computation time. Furthermore, we validate sim-to-real transfer by successfully deploying a policy trained in simulation to real-world mobile manipulation. The code, simulation environments, and pre-trained models will be made available on the project webpage\footnote{https://vbmnet.github.io}.   

\end{abstract}

\section{INTRODUCTION}
Mobile manipulators extend the range of pick-and-place tasks beyond the immediate workspace of the manipulator using their mobility \cite{roa2021mobile}. These tasks are commonly solved in a sequential manner, i.e., the robot first navigates to a suitable base pose for picking/placing and then picks/places the object using a manipulator \cite{reister2022combining}. While some recent works have used whole-body motion without explicitly planning the base poses \cite{haviland2022holistic,sundaresan2025homer}, they only consider last-mile manipulation. In this work, we look into the former approach.

Traditional methods for determining base poses for grasping rely on estimating the 6D pose of the target object, projecting Inverse Reachability Maps (IRM) \cite{vahrenkamp2013robot,makhal2018reuleaux} onto the floor, and selecting feasible base poses based on a model of the environment \cite{reister2022combining}. Recent works have explored data-driven methods that learn base pose selection policies from data \cite{jauhri_robot_2022,naik2024basenet}. However, both traditional and data-driven methods assume a reliable object pose estimate is available, which is often only the case when the object is within the robot's close-range field of view. This limitation delays planning and can result in inefficient behavior due to the repeated need for repositioning of the mobile base \cite{reister2022combining}.

In recent years, visual manipulation -- manipulation using only visual features without requiring precise state estimation -- has gained prominence \cite{zeng2021transporter}. Similarly, in this work, we investigate determining the optimal base pose for manipulation only using visual features, without assuming that the object state and the environment model are available. 

Determining the optimal base pose for manipulation requires an overview of the entire scene. Onboard robot cameras provide an egocentric view, which lacks this global perspective. External cameras placed in the environment can provide the required overview \cite{naik2022multi}; however, the captured scene appearance can vary significantly due to scale changes or distortions introduced by camera viewpoints and lens properties, making it challenging to learn robust visual features. Consequently, extensive data augmentation is often required to achieve generalization. Recent works on tabletop planar pick-and-place tasks have shown that orthographic representations—such as top-down views—can facilitate sample-efficient learning by providing a consistent, scale-preserving scene layout \cite{zeng2021transporter,wang2023robot}.

\begin{figure}[t]
    \centering
    \includegraphics[width=0.48\textwidth]{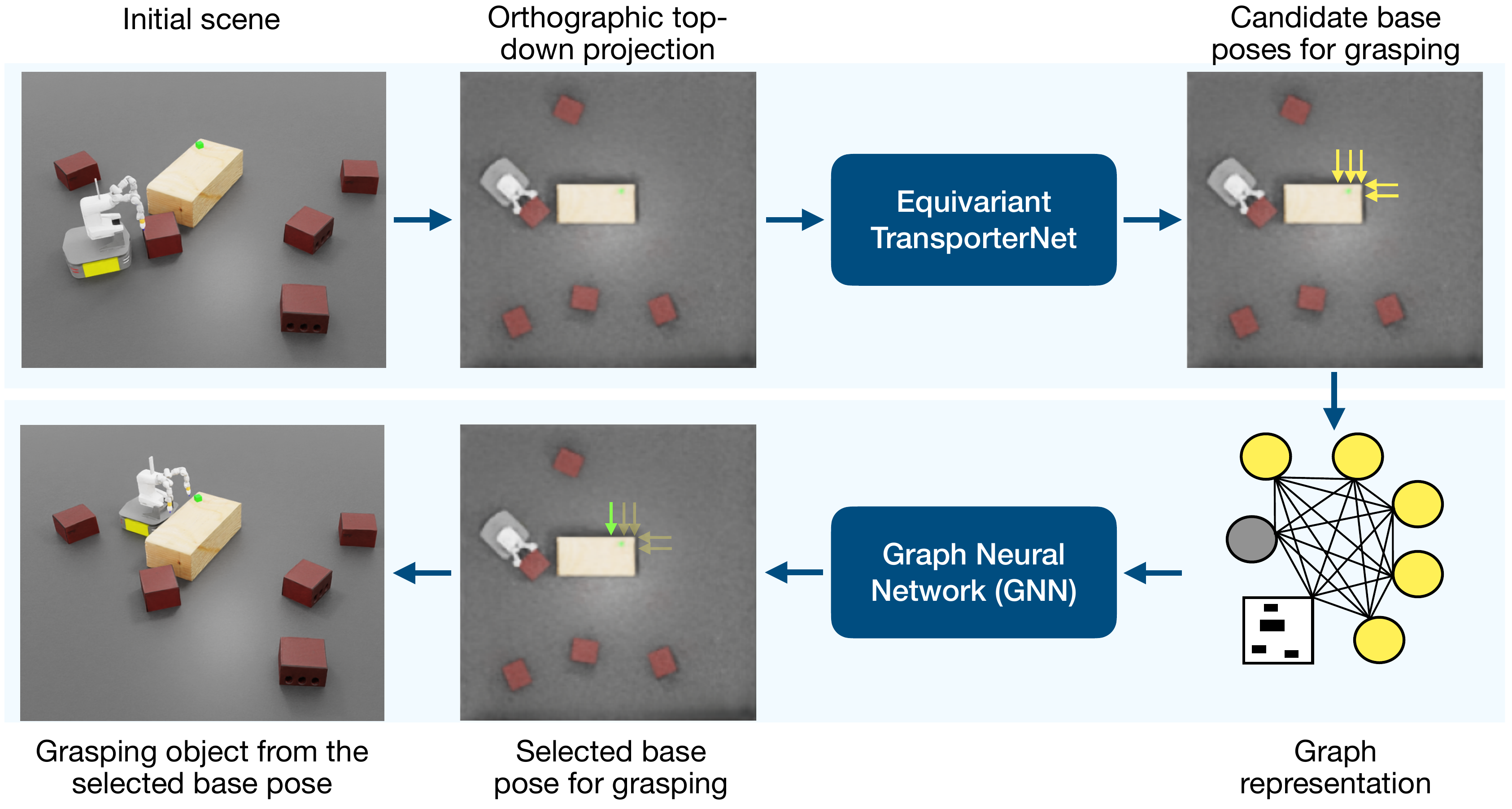}
    \caption{\name~overview. \textbf{Top row:} Orthographic top-down projection of the scene is processed by TransporterNet to predict potential base poses for grasping (yellow arrows). \textbf{Bottom row:} The scene and candidate base poses are encoded as a graph, and a Graph Neural Network is used for selecting the optimal base pose (green arrow).}
    \label{fig:idea}
\end{figure}

Drawing inspiration from these works, we use orthographic scene representations to learn a policy for determining the optimal base pose for manipulating an object using a mobile manipulator. In the context of planar pick-and-place tasks, determining the base pose for manipulation can be interpreted as repositioning the robot from its current base pose to a new base pose that enables successful object manipulation, minimizing the total navigation cost.

However, in contrast to the planar pick-and-place tasks, where only a single valid solution exists for placing \cite{zeng2021transporter}, optimal base pose selection is inherently more complex. An object can often be grasped from multiple base poses with valid Inverse Kinematic (IK) solutions and no collisions with obstacles in the environment. Consequently, the set of potential base poses varies across scenes. Further, base pose planning requires a global overview of the environment, which demands higher-resolution visual features than those used in planar pick-and-place tasks. These factors introduce two key challenges:
\begin{itemize}
    \item \textbf{Sample inefficiency.} Learning in a high-dimensional visual state and spatial action space requires a large amount of training data.
    \item \textbf{Varying number of base poses.} Since the number of feasible base poses varies by scene, they cannot be encoded as a fixed-dimensional vector for learning.
\end{itemize}
To address these challenges, we propose a two-stage approach. First, we use equivariant models \cite{zeng2021transporter,zhu2022sample,huang2022equivariant,huang2024leveraging} to exploit the spatial structure of orthographic top-down projections, improving sample efficiency. In this stage, we train a model to identify potential base poses for grasping in a supervised manner using data derived from IRMs (Fig.~\ref{fig:idea}, top row). Second, to handle the variable number of base poses, we represent them as nodes in a graph and formulate optimal base pose selection as a graph node regression problem (Fig.~\ref{fig:idea}, bottom row). To summarize, we make the following contributions:
\begin{itemize}
    \item We formulate the problem of determining the optimal base pose for grasping the object from visual representations of the scene as a learning problem.
    \item We propose \name, a two-stage approach that first learns potential base poses and then selects the optimal base pose while accounting for navigation cost.
    \item We address sample-efficiency using equivariant models and orthographic projections and a varying number of potential base poses using Graph Neural Networks (GNN's).
    \item Through experimental evaluation, we show that the approach can produce comparable results to classical state-based methods that use ground-truth states.
    \item We validate sim-to-real transfer by deploying the policy trained in simulation in real world.
\end{itemize}


\section{RELATED WORK}
\subsection{Mobile Manipulation Using Visual Representations}
Recent works have focused on learning whole-body motion control directly from visual inputs obtained from the robot's onboard cameras \cite{sundaresan2025homer, yan2025m, zhang2024gamma}. Sundaresan et al. \cite{sundaresan2025homer} has presented an imitation learning framework with hybrid action modes that handle both long-range navigation  and fine-grained manipulation using RGB images and point clouds as input. Yan et al. \cite{yan2025m} has proposed a diffusion-based model that generates whole-body motion trajectories from robot-centric 3D scans. Zhang et al. \cite{zhang2024gamma} has introduced a reinforcement learning approach for a graspability-aware observe-to-grasp policy using RGB-D images. 

These works use visual representations from the robots' onboard cameras that lack a global perspective of the environment. As a result, they focus primarily on last-mile manipulation and ignore the navigation cost for manipulation from the robot’s initial base pose. In contrast, in this work, we use orthographic top-down projections obtained from external cameras, which provide a global overview of the scene and enable explicit reasoning about navigation cost.

\subsection{Explicit base pose planning}
These works involve first planning an explicit base pose for grasping, navigating to the planned base pose, and grasping the object only using the manipulator motion \cite{naik2024basenet, naik2024pre, reister2022combining}. They can be further classified as follows.\\
\textbf{State-based representations.} They use object-centric representations either to plan \cite{reister2022combining} or learn base poses for manipulation \cite{naik2024basenet}. They consider navigation costs.\\
\textbf{Visual representations.} These works learn using visual representations from the robots' onboard camera \cite{wu2025momanipvla, hou2024tamma}. Hou et al. \cite{hou2024tamma} obtain a 3D Gaussian initialization of the whole scene and use it to determine the approach pose. However, they do not explicitly consider the navigation cost to this pose. Wu et al. \cite{wu2025momanipvla} have proposed bi-level optimization framework wherein they predict waypoints for the base to enhance the manipulator policy space without considering any navigation cost. They ignore navigation costs due to a lack of a global overview in the ego-centric visual representations provided by the robots' onboard cameras.

In this work, we plan an explicit base pose for grasping using orthographic top-down projections of the scene while considering the navigation cost. To the best of our knowledge, this is the first work to consider explicit base pose learning from visual representations, considering the navigation cost.

\begin{figure*}[h]
    \centering
    \includegraphics[width=17.5cm]{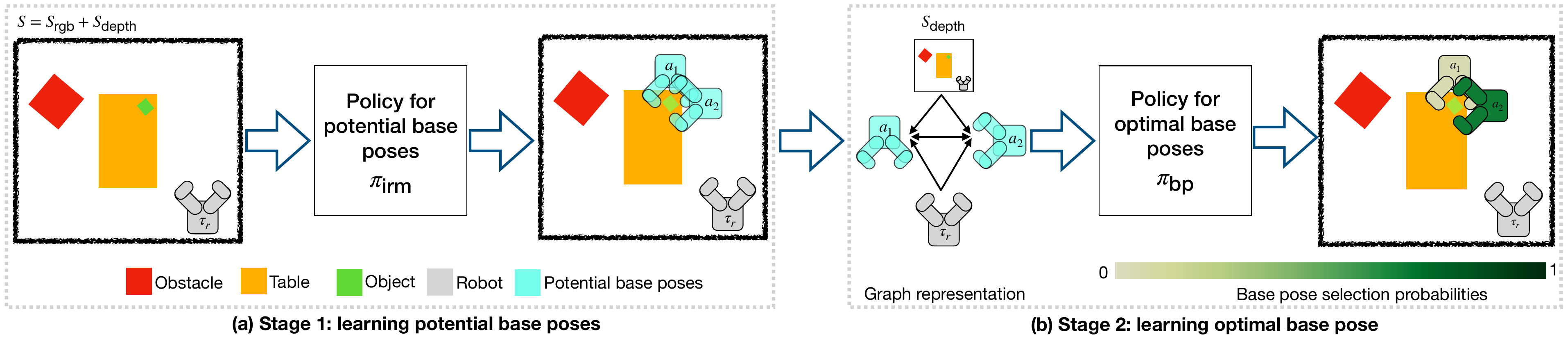}
    \caption{\name~(a) Stage 1: Learning potential base poses for grasping using the policy $\policybp$. (b) Stage 2: Selecting the optimal base pose by incorporating navigation cost, using the policy $\policyobp$, from the candidate poses generated in Stage 1.}
    \label{fig:arch}
\end{figure*}

\subsection{Equivariant Neural Networks (ENN)}
ENNs consist of layers that encode a function 
$f:X\rightarrow Y$ constrained by the \emph{equivariance property}:
\begin{equation}
    g f(x) = f(gx), \quad \forall g \in G,
\end{equation}
where $G$ denotes a finite group. Base poses for grasping exhibits spatial equivariance in SE(2). 
Convolutional layers are naturally equivariant to translations. 
Therefore, in this work, we are primarily interested in equivariance with respect 
to planar rotations, represented by the group $\mathrm{SO}(2)$. 

In practice, to ensure computational tractability, models typically use 
a cyclic subgroup $C_n \subset \mathrm{SO}(2)$:
\begin{equation}
    C_n = \left\{ \tfrac{2\pi k}{n} \;:\; 0 \leq k < n \right\},
\end{equation}
which corresponds to discrete rotations by multiples of $2\pi/n$ radians.  

An equivariant convolutional layer maps between feature maps that transform 
according to specified group representations. 
For a standard convolutional layer, the feature map is a tensor 
$F \in \mathbb{R}^{m \times h \times w}$, 
where $m$ denotes the channel dimension and $h \times w$ the spatial resolution. 
An equivariant convolutional layer introduces an additional 
group dimension, resulting $F \in \mathbb{R}^{n \times m \times h \times w},$ where $n$ corresponds to the dimension of the chosen group representation.

\subsection{Graph Neural Networks (GNN)}
GNNs are used for learning over graph-structured data, where nodes represent entities and edges encode 
their relationships \cite{wu2020comprehensive}. They have been also widely used for robotics applications. 
In this work, we use the Graph Attention Layers \cite{velickovic2017graph} to learn a vector representation that encodes relevant grasp scene information to determine optimal base poses for grasping while considering the navigation costs to the potential base poses.





\section{PROBLEM FORMULATION}


We address the problem of grasping a selected object from a table using a mobile manipulator. We formulate this as a contextual bandit problem in visual state space and spatial action space.

We assume that an orthographic top-down projection of a scene is available \cite{zeng2021transporter,chen2024ortho} and that the robot has a navigation stack \cite{guimaraes2016ros} to navigate to the planned base pose and a manipulation stack \cite{chitta2012moveit} for grasping. Further, we assume that the robot's kinematic model is available and the object can be grasped using a top-down grasp pose.

Often, the object can be grasped from multiple base poses with valid Inverse Kinematics (IK) solutions for grasping. Our goal is to find the optimal base pose with a valid IK solution for grasping the object that minimizes the total navigation cost.

Learning such a policy requires information about the object, the robot, and the obstacles in the scene, including the table with the object to grasp (workspace). All this information is present in the visual orthographic top-down projection of the scene. So the state is represented as
\begin{equation}
    \state = \{ \statergb, \statedepth  \} \in \statespace = \R^{M \times H \times W},
\end{equation}
where $\statergb \in \R^{3 \times H \times W}$ is an orthographic RGB image and $\statedepth \in \R^{1 \times H \times W}$ is the orthographic depth image.

To simplify the learning problem we discretize continuous action space $\actionspace = \setwo$. Thus, action is parameterized as 
\begin{equation}
    \action = \{ \imgpixx, \imgpixy, \theta \} \in \actionspacedis = \R^{K \times H \times W},
\end{equation}
where $(\imgpixx, \imgpixy)$ denote pixel coordinates (in $\state$ corresponding to possible robot base positions), and $\theta$ represents the orientation of the robot, discretized into $K$ uniform angular bins.

Thus, we learn a policy $\policy$ that maps the visual orthographic representation of the scene $\state \in \statespace$ to a base pose $\action \in \actionspacedis$  from where the object can be grasped with a minimal navigation cost.



\section{\name}

In this section, we present \name, a two-stage approach for determining the optimal base pose for grasping, given the orthographic top-down projection of the scene $\state$ and the robot’s kinematic model. In the first stage, we learn a sub-policy $\policybp$ that identifies potential base poses for grasping the object, ensuring both IK feasibility and collision avoidance with the table or other obstacles (Section~\ref{ss:lpbp}; Fig.~\ref{fig:arch}a). In the second stage, we learn a policy $\policyobp$ that selects the optimal base pose from the candidates suggested by $\policybp$, considering the navigation cost (Section~\ref{ss:obp}; Fig.~\ref{fig:arch}b). The overall policy is thus expressed as the composition $\policy = \policybp \times \policyobp$.

\subsection{Stage 1: learning potential base poses}
\label{ss:lpbp}
\textbf{Inverse Reachability Map policy $\policybp$.} Let $\actionvalid(\state)$ be the set of base poses $\action \in \actionspacedis$ for grasping given the top-down orthographic projection of the scene $\state$ from where there exists a valid IK solution for grasping the object $\ik(\state, \action)$ and the robot can exist at the base pose $\action$ without any collision with obstacles in the environment $\neg\collision(\state, \action)$
\begin{equation}
    \actionvalid(\state) = \{ \action \mid \neg\collision(\state, \action) \;\land\; \ik(\state, \action) \}
\end{equation}

We learn the policy $\policybp$ that maps the top-down orthographic projection of the scene $\state$ to the set of valid base poses for grasping $\actionvalid(\state)$ (Fig.~\ref{fig:arch}a)
\begin{equation}
    \policybp : \state \mapsto \actionvalid(\state).
\end{equation}


\begin{figure*}[h]
    \centering
    \includegraphics[width=17.5cm]{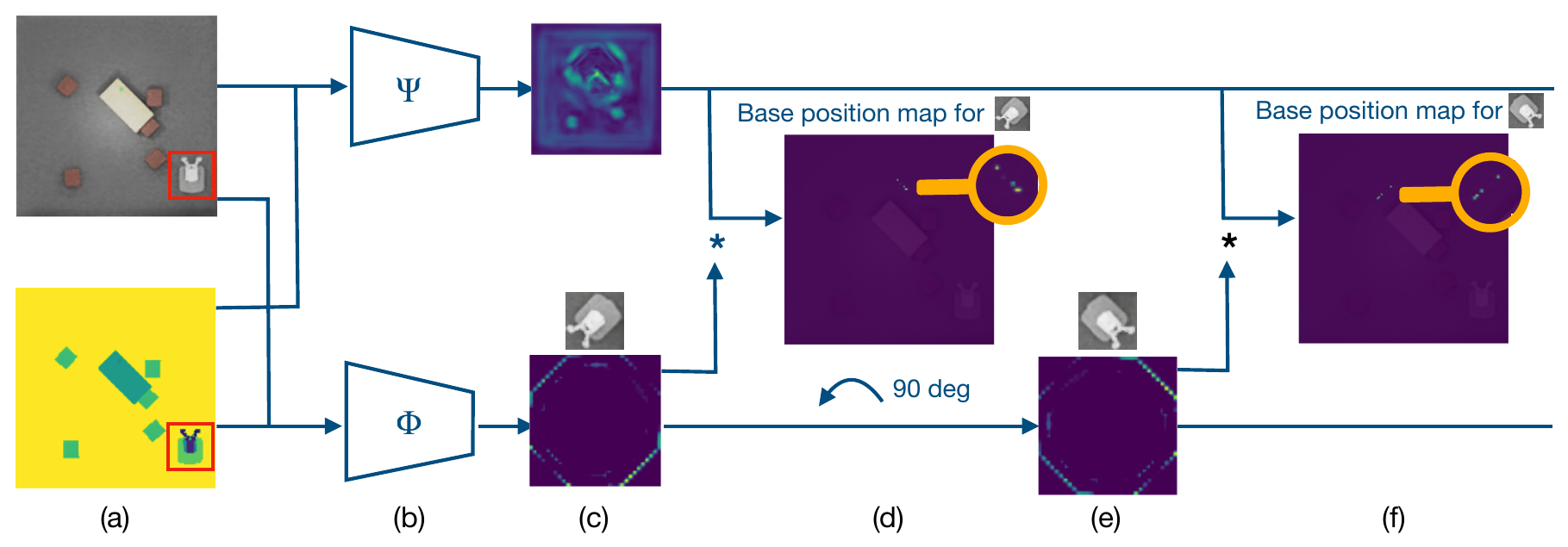}
    \caption{Learning potential base poses for grasping the object: \textbf{(a)} Input RGB and depth images of the scene with the cropped robot query. \textbf{(b)} Equivariant networks for extracting feature embeddings with TransporterNet. \textbf{(c)} Generated feature embeddings. \textbf{(d)} Identified feasible base positions with the robot’s original orientation + 135$\degree$ (counterclockwise). \textbf{(e)} Rotated robot feature embeddings. \textbf{(f)} Identified feasible base positions with the robot’s original orientation + 225$\degree$ (counterclockwise).}
    \label{fig:tnsp_arch}
\end{figure*}

\textbf{Learning problem.} We formulate this as a multi-label, pixel-wise classification problem, where the goal is to predict the feasibility of placing the robot base at various orientations $\theta_k$ for each pixel location $(\imgpixx, \imgpixy)$ in the orthographic top-down image $\state$. Accordingly, the output of the policy $\policybp$ is a stack of $K$ density maps, $\policybp(\state) \in [0,1]^{K \times H \times W}$, where each channel corresponds to a discrete orientation $\theta_k$ and predicts the probability of selecting $(\imgpixx, \imgpixy, \theta_k)$ as a feasible base pose. It is trained in a supervised manner using a dataset of $P$ expert demonstrations $\dataset = \{\state_i, \action_i \}_{i=1}^{i=P} $, obtained using the IRM \cite{vahrenkamp2013robot,makhal2018reuleaux}.

\textbf{Learning $\policybp$ via transporting.} The distribution of potential base poses for grasping over pixels in $\state$ is inherently multi-modal. Furthermore, the problem exhibits spatial equivariance in SE(2), i.e., if the state is rotated by group action $g$, then the potential base poses transform accordingly
\begin{equation}
     \policybp : g \cdot \state \mapsto g \cdot \actionvalid(\state).
\end{equation} 
As a result, we learn $\policybp$ using the Equivariant TransporterNet \cite{huang2024leveraging}, which can represent complex multi-modal distributions while preserving spatial equivariance in SE(2).

In TransporterNet, the goal is to transport a crop of an object densely across a set of poses to search for the placement with the highest feature correlation. In this work, we transport the crop of the robot densely over a set of candidate base poses $\action \in \actionspacedis$ to search for valid base poses for grasping $\actionvalid(\state)$. If $\robotpos = (u_r, v_r)$ is the robot center in $\state$, then $\state[\robotpos]$, which is the robot crop is referred to as the \textit{query} (Fig.~\ref{fig:tnsp_arch}a, red box) and $\state$, which is a search space, is referred to as the \textit{key} (Fig.~\ref{fig:tnsp_arch}a). Thus, determining potential base poses for grasping $\action$ is formulated as a template matching problem, using cross-correlation (Fig.~\ref{fig:tnsp_arch}d, f) with dense feature embeddings $\Psi(\state[\robotpos])$ and $\Phi(\state)$ (Fig.~\ref{fig:tnsp_arch}c, e) from two deep equivariant models:
\begin{equation}
    Q_{\text{irm}}(\action | \state, \robotpos) = \Psi(\state[\robotpos]) * \Phi(\state)[a].
\end{equation}

In contrast to TransporterNets, where the features to correlate are directly located at the candidate grasp pose (i.e., the object is physically placed at the queried pose), in our problem, not all relevant features necessarily lie at the candidate base pose itself. For example, features related to collision checking are local to the robot base pose, whereas features that determine whether the target object can be grasped may be located elsewhere in the scene. However, since the deep feature embeddings are spatially overlaid, the influence of the feature correlation extends beyond the immediate crop region \cite{zeng2021transporter}.


\subsection{Stage 2: learning optimal base pose}
\label{ss:obp}
\textbf{Optimal base pose policy $\policyobp$.} The policy $\policyobp$ selects the optimal base pose for grasping, considering the navigation cost among the candidate base poses $\actionvalid(\state)$ proposed by $\policybp$. The likelihood of selecting a particular base pose depends on the current robot pose, the set of other potential base poses, and the navigation costmap. Since the number of candidate base poses varies across scenes, we use a GNN to encode a permutation-invariant, fixed-dimensional context embedding for each base pose in $\actionvalid(\state)$.

\textbf{Encoder.} We represent the problem as a heterogeneous graph with three types of nodes: the robot node $\robot$, a candidate base pose node under consideration $\action_i$, and nodes for other potential base poses $\action_j$. In addition, the orthographic depth projection of the scene $\statedepth$ is encoded using a ResNet \cite{he2016deep} to provide a global navigation costmap.  

\begin{figure}[h]
    \centering
    \includegraphics[width=0.48\textwidth]{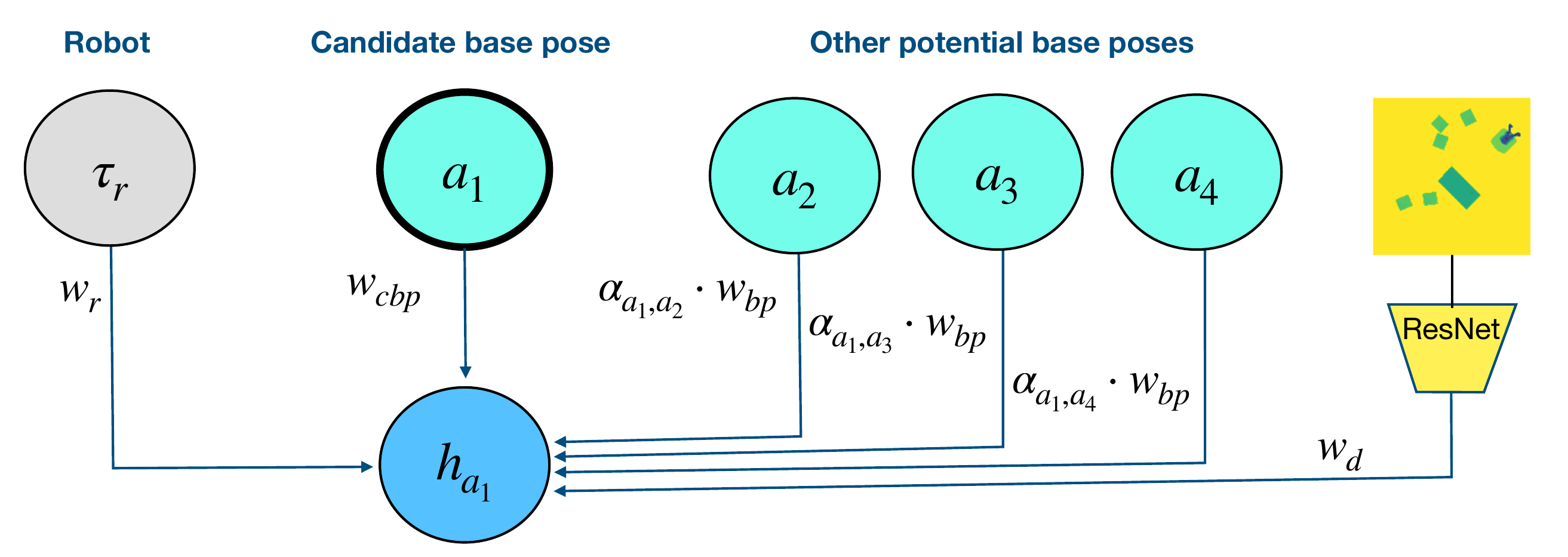}
    \caption{Attention-based graph encoder for candidate base poses in $\actionvalid(\state)$.}
    \label{fig:encoder}
\end{figure}

We use Graph Attention Layers \cite{velivckovic2018graph} to encode relevant information into a context embedding. Each node is first projected into a shared latent space using learnable weights: $w_r$ for the robot position $\robotpos$, $w_{\text{cbp}}$ for the candidate base pose $\action_i$, $w_{\text{bp}}$ for the other base poses $\action_j$, and $w_{d}$ for the costmap embedding. Attention coefficients $\alpha$ are then computed to weigh the contribution of other base poses $\action_j$ when encoding the context embedding of the candidate $\action_i$. The context embedding for $\action_i$ is thus given by
\begin{equation}
    h_{\action_i} = w_{\text{cbp}} \cdot \action_i + w_r \cdot \robotpos + w_{d} \cdot \mu(\statedepth) + \sum_{j \in \Omega(\action_i)}  \alpha_{\action_i, \action_j} \cdot w_{\text{bp}} \cdot \action_j,
\end{equation}
where $\Omega(\action_i) = \actionvalid(\state) \setminus \{ \action_i \}$ and $\mu(\statedepth)$ denotes the ResNet embedding of $\statedepth$. 
The encoder architecture is illustrated in Fig.~\ref{fig:encoder}.

\textbf{Learning problem.} We formulate the optimal base pose policy $\policyobp$ learning as a graph node regression problem. The selection likelihood of each base pose is computed as
\begin{align}
y_n &= \policyobp(\ \cdot\ | \{ h_{\action_n} \}_n ; \phi_{\text{obp}}), \quad n \in \{1...N\}, \nonumber \\
k &= \argmax_{n} \{ y_n \}_n,
\end{align}
where $y_n$ denotes the likelihood of selecting the $n$-th candidate base pose, $h_{\action_n}$ is its context embedding, $N = |\actionvalid(\state)|$ is the number of candidates, $\policyobp$ is the policy parameterized by $\phi_{\text{obp}}$, and $k$ indexes the selected base pose with the highest probability.

\textbf{Learning $\policyobp$ via Reinforcement Learning.} The policy $\policyobp$ is trained using REINFORCE \cite{sutton2018reinforcement} with a greedy baseline. The baseline corresponds to the candidate with the lowest navigation cost among those proposed by $\policybp$. By restricting the decision space to $\actionvalid(\state)$ (instead of $\actionspacedis$), the exploration complexity is significantly reduced. The reward for selecting a base pose $\rewardbp$ is defined as
\begin{align}
\rewardbp = \;& \rparam{1} \cdot \mathds{1}\!\left(\neg\,\collision(\action_n)\right) 
+ \rparam{2} \cdot \mathds{1}\!\left(\ik(\action_n)\right) \nonumber \\
&+ \rparam{3} \cdot \nav(\action_n).
\end{align}
where $\mathds{1}(\neg\;\collision(\action_n)) = 1$ if the robot does not collide with the environment at $\action_n$, $\mathds{1}(\ik(\action_n)) = 1$ if a valid IK solution exists and $\nav(\action_n)$ is the navigation cost from the current robot pose to $\action_n$. The coefficients $\rparam{1}$, $\rparam{2}$, and $\rparam{3}$ are hyperparameters.  

Each training episode consists of a single decision step. The encoder produces fixed-dimensional embeddings $h_{\action_n}$  for all candidates $\{ \action_n \}_{n=1}^N$, which are passed to an MLP decoder that predicts selection likelihoods. The base pose is chosen by sampling from a categorical distribution parameterized by these likelihoods. The policy gradient is then computed as
\begin{equation}
    \nabla \mathcal{L}(\phi_{\text{obp}}|\state) = - \big( \rewardbp(\state, \action_n) - \rbaseline(\state) \big) \cdot \nabla \log y_n,
\end{equation}
where $\rbaseline(\state)$ is the baseline reward used to reduce variance. It is computed similar to the $\rewardbp$. Since infeasible navigation paths can lead to large reward differences, which destabilize training, we apply logarithmic scaling to the advantage term:
\begin{equation}
    A' = \log \big(1 + \max(\epsilon, \rewardbp(\state, \action_n) - \rbaseline(\state)) \big),
\end{equation}
where $\epsilon > 0$ ensures numerical stability. This scaling helps stabilize learning by reducing sensitivity to large reward differences.


\section{EXPERIMENTAL SETUP}
\subsection{Experimental setup and implementation details}

\textbf{Simulation environment.} We evaluate our approach in a simulation environment built in NVIDIA Isaac Sim, using a mobile manipulator and a rectangular table (1.6\,m $\times$ 0.8\,m) with a green cube object of size 7.5\,cm, as shown in Fig.~\ref{fig:qual_res}. The grasp pose for the object is predefined as a top-down grasp. Orthographic top-down projections of the scene are generated at a resolution of $160 \times 160$ pixels, with 8 orientation channels for robot base orientations discretized at $45^\circ$ intervals ($C_8$ finite group).

\textbf{Training $\policybp$.} For training the \textit{Inverse Reachability Map} policy $\policybp$, we use an Equivariant Transporter Network \cite{huang2022equivariant}, where both the key and query encoders, $\Phi$ and $\Psi$, are implemented as 4-layer Equivariant U-Nets with channel sizes ${16, 32, 64, 128}$ \cite{zhu2022sample}. The networks are trained jointly in a supervised manner using the Adam optimizer \cite{kingma2014adam} with a learning rate of $1\times10^{-4}$, a batch size of 32, and 
mean squared error (MSE) loss.

To generate training data, the table and obstacles are first placed randomly in the scene, followed by placing the object at a random location on the table. The robot’s initial pose is sampled within a 3m radius around the table, ensuring no collisions with the environment (Fig.~\ref{fig:qual_res}). The orthographic projection of the scene and the robot’s bounding box are obtained directly from the simulator’s internal state.  IRMs are computed using the Lula Kinematics Solver in NVIDIA Isaac Sim with a discretization resolution of 10cm in position and $45^\circ$ in orientation. The final training dataset consists of 15,000 such samples. Training takes approximately one day on an NVIDIA A100 GPU with 40 GB VRAM.

\textbf{Training $\policyobp$.} For training the \textit{optimal base pose} policy $\policyobp$, we use a 64-dimensional graph-based representation learned through three Graph Attention Layers (GATs) \cite{velivckovic2018graph}, each producing a 64-dimensional embedding followed by ReLU activation. The resulting context vector is passed through a multi-layer perceptron (MLP) with two hidden layers of 64 neurons each, also followed by ReLU activations. The orthographic depth projection $\statedepth$ is encoded using a pretrained ResNet backbone, which outputs a 64-dimensional embedding. All components are trained jointly with the Adam optimizer \cite{kingma2014adam} using a learning rate of $1\times 10^{-4}$. Policy gradients are computed using the REINFORCE algorithm \cite{sutton2018reinforcement} with a greedy baseline. The reward weighting coefficients $\rparam{1}$, $\rparam{2}$, and $\rparam{3}$ were set to 1, 1, and -1 (lower the navigation cost better), respectively.

Each training episode is generated in the same manner as the data used for training $\policybp$. To accelerate learning, the robot is teleported directly to the selected base pose instead of using a navigation stack. Navigation cost is computed using the A* planner, and grasp feasibility is verified using the Lula Kinematics Solver. 
The policy converges after approximately 20,000 interactions, requiring around 15 hours of training on a workstation with an Intel Core i9-13900KF, 64 GB RAM, and an NVIDIA Ada 2000 GPU with 16 GB VRAM.

\subsection{Experiment objectives}
The experiments aim to verify whether the proposed learning-based method can produce solutions comparable to those obtained using classical non-learning-based methods in a shorter computation time (Section~\ref{sec:res:mm}). The following baselines are considered:
\begin{itemize}
    \setlength\itemsep{0.15em}
    \item[] \textbf{Fixed base poses (FBP)}: The robot selects the closest pose from a predefined set of base poses. If not successful, it selects the next closest base pose in the set.
    \item[] \textbf{Proximity-Based Selection (PBS)}: IRMs are used to generate feasible base poses for grasping the object. The robot then selects the base pose with the shortest Euclidean distance to its current location.
    \item[] \textbf{Navigation Cost-Based Selection (NBS)}: Similar to the PBS, IRMs are used to compute a set of feasible base poses. However, the robot selects the base pose with the minimum navigation cost.
\end{itemize}
PBS and NBS are state-dependent methods, as they require knowledge of the object pose to compute the IRMs. FBP is a heuristic method that operates independently of the object state.

In addition, we conduct two ablation studies to validate our design choices (Section~\ref{ss:res:ab}). First, we evaluate the learning performance of $\policybp$ with and without the equivariant architecture. Second, we analyze the learning performance of $\policyobp$ using different RL algorithms.



\section{RESULTS}
\subsection{Comparison with Mobile manipulation methods}
\label{sec:res:mm}

\subsubsection{Qualitative results}
\label{ss:qual_res}
\begin{figure}[h]
    \centering
    \includegraphics[width=0.48\textwidth]{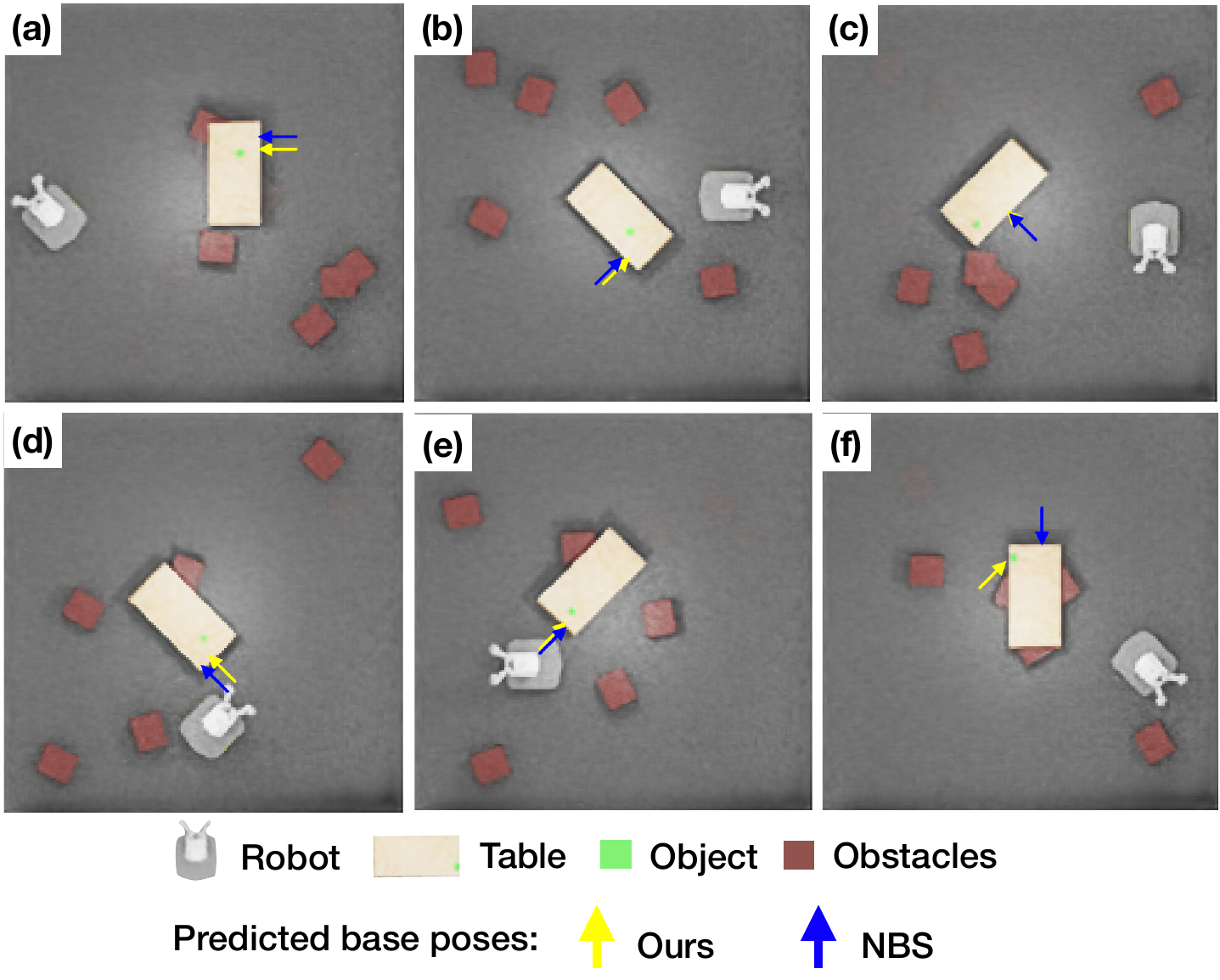}
    \caption{Qualitative results for \name~and NBS. \textit{Note:} The tail of each predicted base-pose arrow corresponds to the robot’s center.}
    \label{fig:qual_res}
\end{figure}

In Fig.~\ref{fig:qual_res}, we present qualitative results for six random scenes using \name~and NBS (best performing baseline in Section~\ref{ss:quant_res}). In most cases (Fig.~\ref{fig:qual_res}a–e), the base poses predicted by \name~are comparable to those predicted by NBS. 

We observed that \name~tends to predict base poses closer to the object (e.g., Fig.~\ref{fig:qual_res}a and d), which results in slightly higher navigation costs compared to NBS. This occurs because $\policybp$ avoids predicting poses farther away from the object, which can risk invalid IK solutions. Consequently, $\policyobp$ is constrained to choose among these base poses, even when other feasible base poses with lower navigation costs exist. This limitation was identified as one source of sub-optimality in minimizing navigation cost.

Another representative failure case is shown in Fig.~\ref{fig:qual_res}f, where \name~predicts a base pose too close to an obstacle, preventing successful execution. We expect performance to improve with finer base-pose discretization, compared to the current setting where each pixel in the orthographic projection corresponds to approximately 3cm and robot orientations are discretized into 8 channels at $45^\circ$ intervals.

\subsubsection{Quantitative results}
\label{ss:quant_res}
In Table~\ref{tab:qual-results}, we report the mean and standard deviation of \textit{planning time}, \textit{navigation path length}, and \textit{success rate} for three baseline methods and the proposed approach. Each method was evaluated on 100 randomly generated scenes. The FBP baseline assumes that the table model and environment model are available, and that a fixed set of base poses around the table has been pre-defined. The PBS and NBS baselines assume access to the full state of the scene, including the object pose, table, and environment model. In practice, state estimation time should also be included in the \textit{planning time}, and the \textit{success rate} of these baselines depends directly on the quality of state estimation \cite{naik2024pre}. 
In our experiments, ground-truth states provided by the simulator were used.

\begin{table}[h]
\centering
\resizebox{0.48\textwidth}{!}{%
\begin{tabular}{@{}cccc@{}}
\toprule
Method &
  \begin{tabular}[c]{@{}c@{}}Planning time \\ (in seconds)\end{tabular} &
  \begin{tabular}[c]{@{}c@{}}Navigation path length\\ (in distance units)\end{tabular} &
  \begin{tabular}[c]{@{}c@{}}Success rate\\ (in \%)\end{tabular} \\ \midrule
FBP  & 02.58$\pm$3.46 & 106.85$\pm$96.44 & 100.0 \\
PBS  & 04.85$\pm$4.40 & $\phantom{0}$67.72$\pm$36.71 & 100.0 \\
NBS  & 05.98$\pm$3.35 & $\phantom{0}$\textit{56.60}$\pm$\textit{36.73} & 100.0 \\ \midrule
\name & \textit{00.14}$\pm$\textit{0.06} & $\phantom{0}$61.68$\pm$37.61 & $\phantom{0}$81.3 \\ \bottomrule
\end{tabular}%
}
\caption{Planning time, navigation path length, and success rate comparison with different mobile manipulation baselines.}
\label{tab:qual-results}
\end{table}

Scenes with no valid base poses (e.g., when randomly placed obstacles blocked all poses with valid IK solutions according to the IRM) were excluded from evaluation. As a result, all baselines achieved a $100\%$ success rate. \name~achieves the lowest planning time, while navigation path lengths are slightly longer than those of NBS, as discussed in Section\ref{ss:qual_res}. The success rate (number of times base pose with minimum navigation cost is selected) of \name~is slightly above $80\%$, with most failures arising due to the reasons discussed in Section\ref{ss:qual_res}.


\subsection{Ablation studies} 
\label{ss:res:ab}

\noindent\textbf{Learning performance $\policybp$.} In Table~\ref{tab:ab1}, we report the impact of using the equivariant TransporterNet architecture \cite{huang2022equivariant} for learning the policy $\policybp$. For comparison, we evaluate a \textit{UNet} \cite{ronneberger2015u} with approximately the same number of parameters as TransporterNet. Performance is measured using standard semantic segmentation metrics: IoU, Dice coefficient, Precision, and Recall, on unseen data. Both networks were trained with the same dataset and number of epochs. The results show that \textit{equivariant TransporterNet} outperforms \textit{UNet} across all metrics.    

\begin{table}[h]
\centering
\resizebox{0.48\textwidth}{!}{%
\begin{tabular}{@{}ccccc@{}}
\toprule
Architecture               & IoU   & Dice  & Precision & Recall \\ \midrule
UNet \cite{ronneberger2015u}                      & 0.6612 & 0.6618 & 0.6644     & 0.6613  \\
Equivariant TransporterNet \cite{huang2024leveraging} & \textit{0.8402} & \textit{0.8460} & \textit{0.8537}     & \textit{0.8434}  \\ \bottomrule
\end{tabular}%
}
\caption{IoU, Dice coefficient, Precision and Recall comparison for different architectures for the policy $\policybp$.}
\label{tab:ab1}
\end{table}

\noindent\textbf{Learning performance $\policyobp$.} In this study, we compare the learning performance of the policy $\policyobp$ using several standard RL algorithms: offline methods such as \textit{DQN} \cite{mnih2015human} and \textit{SAC-discrete} \cite{christodoulou2019soft}, as well as online methods such as \textit{REINFORCE} \cite{sutton2018reinforcement}. The problem involves a high-dimensional action space ($8 \times 160 \times 160$), which makes exploration particularly challenging. To address this, we use the actions predicted by $\policybp$ as priors to encourage the agent to pick those actions. These priors are incorporated as Q-priors: for \textit{DQN}, they are used as a restriction on the decision space, masking the networks output to the candidate base poses predicted by $\policybp$, while for \textit{SAC-discrete}, they are added to the critic output. Due to the large state and action spaces, the replay buffer size was limited to 5,000. For \textit{REINFORCE}, similar to \name, we construct a graph from the base poses predicted by $\policybp$ and formulate the problem as a graph node regression task.

\begin{figure}[h]
    \centering
    \includegraphics[width=0.45\textwidth]{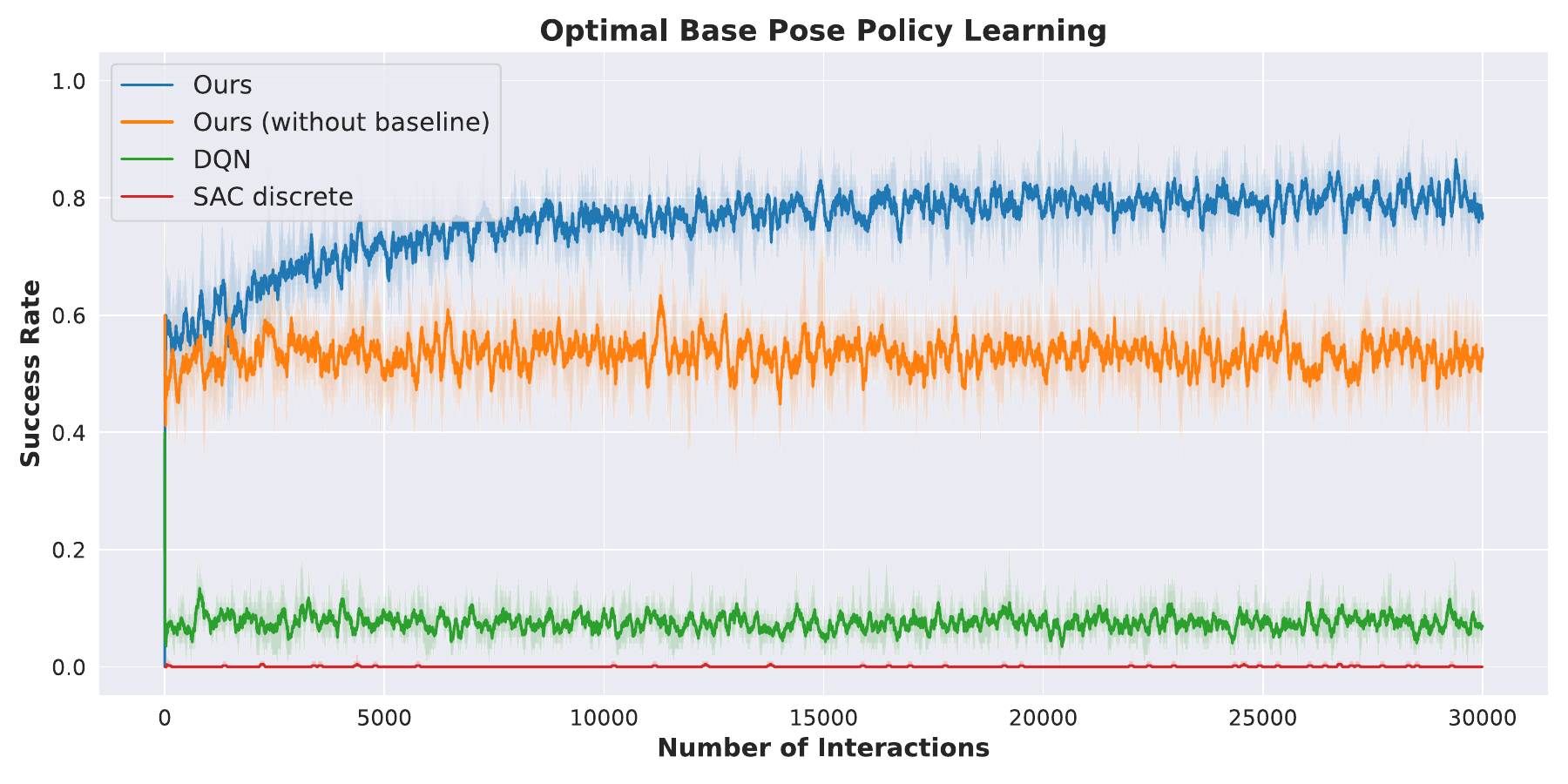}
    \caption{Comparing the learning performance of the optimal base pose policy $\policyobp$ using different RL algorithms. Plot show the mean over 5 seed runs and min and max variations.}
    \label{fig:obp_learning}
\end{figure}

Fig.~\ref{fig:obp_learning} shows the success rate over the last 100 interactions, i.e., the fraction of times the learned policy selected the optimal base pose while considering navigation cost. It can be observed that \textit{DQN} and \textit{SAC-discrete} fail to learn meaningful policies, even with Q-priors. While these methods might benefit from larger replay buffers and more interactions, scaling is infeasible given the high-dimensional state and action spaces. In contrast, \textit{REINFORCE} achieves a success rate of approximately $50\%$ due to its reduced graph state representation based on the candidate base poses suggested by $\policybp$ instead of visual inputs. Our proposed method achieves a success rate of over $80\%$, benefiting from the additional supervision provided by the greedy baseline.


\subsection{Sim-to-real transfer}
\begin{figure*}[ht]
\begin{minipage}[b]{0.30\linewidth}
\centering
\includegraphics[width=\textwidth]{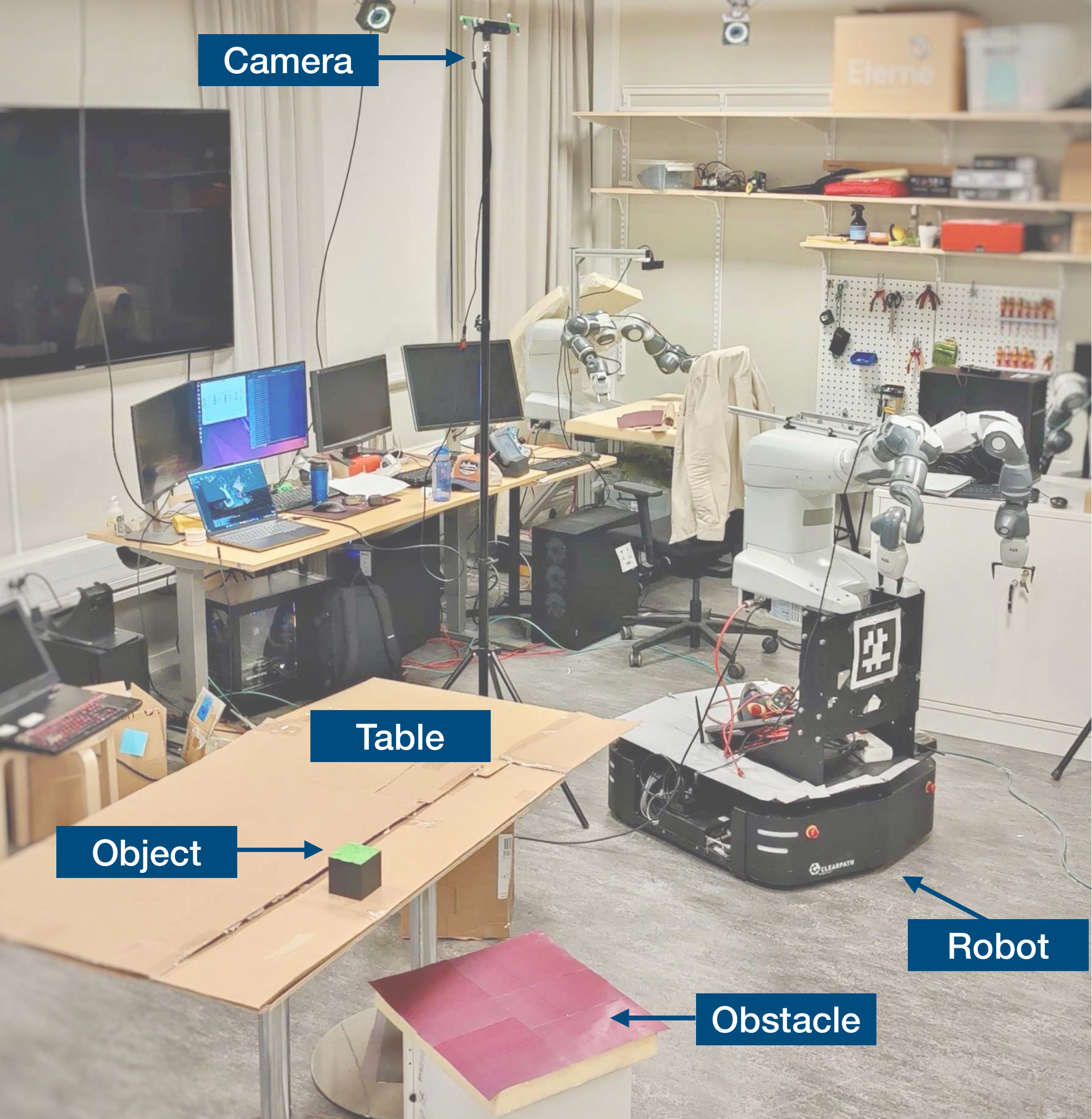}
\subcaption{Real-world experimental setup.}
\label{fig:real_world_setup}
\end{minipage}
\hspace{0.5cm}
\begin{minipage}[b]{0.67\linewidth}
\centering
\includegraphics[width=\textwidth]{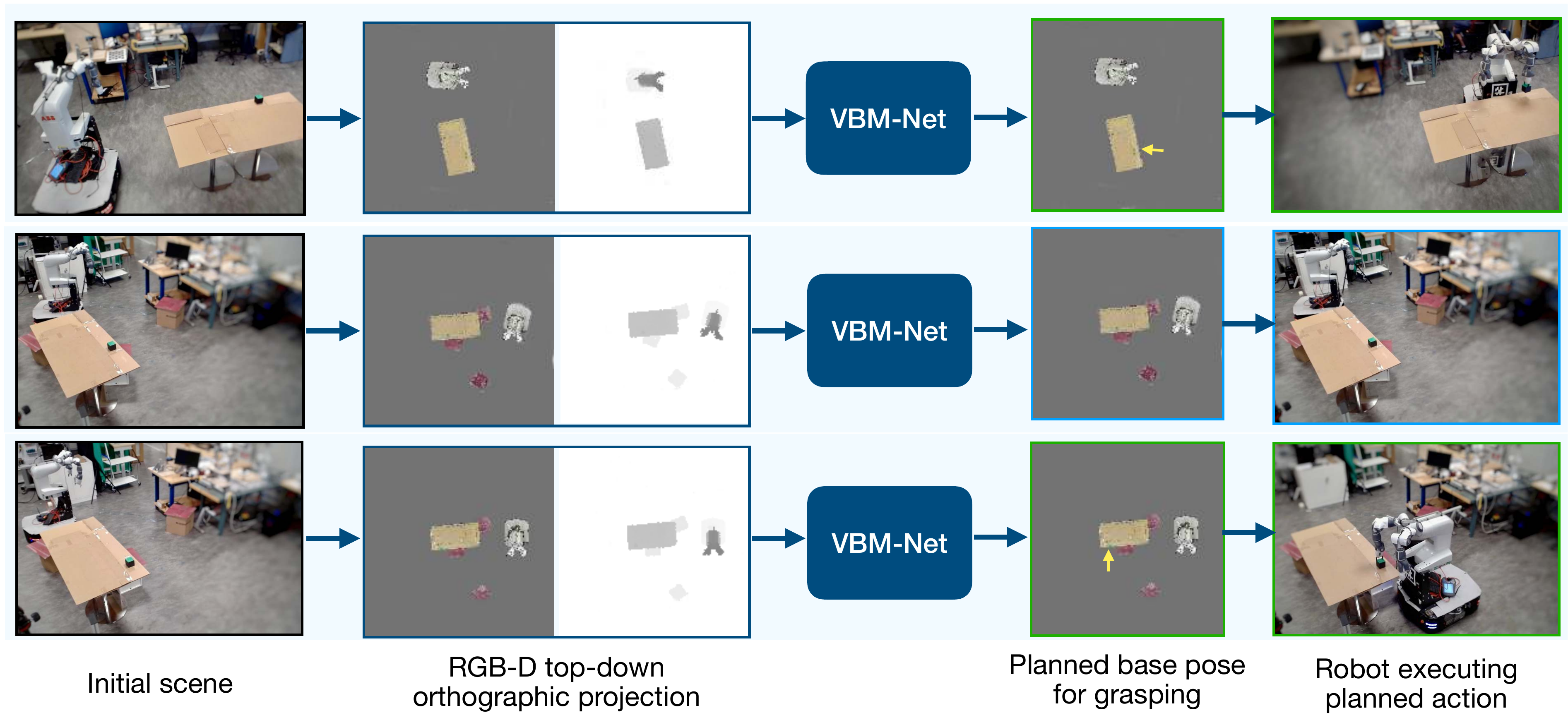}
\subcaption{Sim-to-real transfer examples showing successful base pose prediction \& grasp execution.}
\label{fig:rw_egs}
\end{minipage}
\caption{Real-world demonstrations.}
\label{fig:rwd}
\end{figure*}

For the sim-to-real demonstration, we recreated the training scene in simulation, as shown in Fig.~\ref{fig:real_world_setup}. Orthographic projections of the scene were generated using UFOMap~\cite{duberg2020ufomap}. RGB-D images from 8 external (Orbbec Gemini 2 L) camera views were integrated into UFOMap. The camera poses were estimated using mocap and fine-tuned using Iterative Closest Point~\cite{basl1992icp}. Camera depth errors were handled by incorporating multiple RGB-D images from each camera view (50 per camera view in our experiments). The clutter in the scene was removed from the orthographic projections. Fig.~\ref{fig:rw_egs} presents three representative sim-to-real demonstrations. In the first example (Fig.~\ref{fig:rw_egs}, first row), we consider a simple scene without obstacles and \name~successfully predicts the base pose for grasping. In the second example (Fig.~\ref{fig:rw_egs}, second row), an obstacle is placed near the object, resulting in no valid base pose for grasping without collision with this obstacle. \name~successfully identifies this and does not predict any base pose. In the third example (Fig.~\ref{fig:rw_egs}, third row), the object and obstacle are repositioned, creating valid base poses for grasping on two sides of the table. Here, \name~predicts multiple feasible poses and selects the one with the minimum navigation cost. Full real-world execution videos are provided in the supplementary material.

\section{CONCLUSION AND FUTURE WORK}
In this work, we presented \name, a learning-based method for base pose planning in mobile manipulation. Our experiments demonstrate that it produces solutions comparable to classical methods, using only top-down orthographic projections without requiring precise state estimates. In this way, \name~can enable mobile manipulators to efficiently approach the objects for grasping while considering the navigation cost just using visual representations.

\name~also has several limitations. First, it was trained with a specific type of table, object, and obstacles, and therefore does not generalize well to novel scenes. This limitation could be addressed by incorporating semantic segmentation masks to improve scene understanding. Second, the method assumes planar grasps, i.e., objects can be grasped using top-down grasps. This assumption may not hold for all objects, and extending the approach to 3D orthographic projections would be necessary. Finally, the base poses predicted by \name~are limited by quantization errors introduced by discretization. In future work, we plan to extend \name~to plan a sequence of base poses for grasping multiple objects.







\bibliographystyle{IEEEtran}
\typeout{}
\bibliography{references}

\end{document}